# Distance-Geometric Graph Attention Network (DG-GAT) for 3D Molecular Geometry

Daniel T. Chang (张遵)

*IBM (Retired)* dtchang43@gmail.com

**Abstract:** Deep learning for molecular science has so far mainly focused on 2D molecular graphs. Recently, however, there has been work to extend it to 3D molecular geometry, due to its scientific significance and critical importance in real-world applications. The 3D distance-geometric graph representation (DG-GR) adopts a unified scheme (distance) for representing the geometry of 3D graphs. It is invariant to rotation and translation of the graph, and it reflects pair-wise node interactions and their generally local nature, particularly relevant for 3D molecular geometry. To facilitate the incorporation of 3D molecular geometry in deep learning for molecular science, we adopt the new graph attention network with dynamic attention (GATv2) for use with DG-GR and propose the 3D distance-geometric graph attention network (DG-GAT). GATv2 is a great fit for DG-GR since the attention can vary by node and by distance between nodes. Experimental results of DG-GAT for the ESOL and FreeSolv datasets show major improvement (31% and 38%, respectively) over those of the standard graph convolution network based on 2D molecular graphs. The same is true for the QM9 dataset. Our work demonstrates the utility and value of DG-GAT for deep learning based on 3D molecular geometry.

## 1 Introduction

Deep learning for molecular science has so far mainly focused on *2D molecular graphs*. Recently, however, there has been work to extend it to *3D molecular geometry*, due to its scientific significance and critical importance in real-world applications [6].

*3D Molecular geometry* is commonly represented as *3D graphs*. As in [2-3], we focus on 3D graphs whose geometry can be fully specified in terms of *edge distances (d), angles (θ)* and *dihedrals (φ)* (a.k.a. torsion angles). A key advantage of such specification is its *invariance to rotation and translation* of the graph.

*Distance geometry* [2-5] is the characterization and study of the geometry of 3D graphs based only on given values of the *distances* between pairs of nodes. From the perspective of distance geometry, the geometry of 3D graphs can be equivalently specified in terms of *edge distances (d), angle distances ($d^θ$)* and *dihedral distances ($d^φ$)*. In addition to *invariance to rotation and translation* of the graph, such specification adopts a *unified scheme (distance)* and reflects *pair-wise node interactions* and their generally local nature.

To facilitate the incorporation of geometry in deep learning on 3D graphs, three types of *geometric graph representations* are defined in [2]: positional, angle-geometric and distance-geometric. The *positional graph representation* is based on node positions, i.e., Cartesian coordinates of nodes. The *angle-geometric graph representation* centers on edge distances (d), angles (θ) and dihedrals (φ); it is invariant to rotation and translation of the graph. The *distance-geometric graph representation (DG-GR)* is based on *distances*: edge distances (d), angle distances ($d^θ$) and dihedral distances ($d^φ$); it is invariant to rotation and translation of the graph, and it reflects pair-wise node interactions and their generally local nature, particularly relevant for 3D molecular geometry. As in [2-3], we focus on *DG-GR* since it adopts a *unified scheme (distance)* for representing the geometry of 3D graphs.

*Graph convolutional networks (GCNs)* [7] have been applied to deep learning on graphs. However, standard GCNs do not take spatial arrangements of the nodes and edges into account. Therefore, they can accommodate only 2D graph constitution, but not 3D graph geometry. To incorporate geometry in graph convolutions, *geometric graph convolutions* [2] use DG-GR and employ parameterized edge weight / edge distance power laws. The combination enables the incorporation of geometry in graph convolutions utilizing standard GCNs by (1) expanding the kinds of edges involved to include not just *(connected) edges (e)* with neighbor nodes, but also *angle edges ($e^θ$)* with second-order-neighbor nodes and *dihedral edges ($e^φ$)* with third-order-neighbor nodes, and (2) assigning different *weight*s to different edges based on their kind and distance.

*DG-GCN (distance-geometric graph convolutional network)* [3], a message-passing graph convolutional network based on DG-GR, improves upon this. Similar to geometric graph convolutions, DG-GCN considers all edges that are important to 3D graph geometry in graph convolutions. These include *(connected) edges*, *angle edges* and *dihedral edges*. However, instead of standard GCN layers, it utilizes *continuous-filter convolutional layers (CFConv)*, with *filter-generating networks*, which enable learning of filter weights from edge distances. This avoids the need for hand-crafted edge weight / edge distance power laws.

In this work, to make further improvement, we adopt the new *graph attention network with dynamic attention (GATv2)* [8-9] for use with DG-GR and propose the 3D *distance-geometric graph attention network (DG-GAT)*. Essentially, we replace CFConv in DG-GCN (or standard GCN in geometric graph convolutions) with GATv2 as the convolution layer in DG-GAT. GATv2 is a great fit for DG-GR since the attention can vary by node and by distance between nodes. Further, GATv2 is simpler to understand and use than CFConv, and it is more general and more powerful.

DG-GAT is implemented using *PyTorch Geometric (PyG)* [10]. In particular, the implementation adopts *GATv2Conv* as the convolutional layer, with the attention coefficients computed based on both *node features and edge features*. For experiments, we use the *ESOL* and *FreeSolv* datasets provided by geo-GCN [11] as well as the *QM9* dataset in PyG, which contain molecular graph data including *3D node coordinates*.

# 2 Background Information

## 2.1 2D Molecular Graph

A large fraction of molecules can be treated as *2D molecular graphs* [6], at least for certain applications: the *atoms* and *bonds* are nodes and edges in the graph, respectively. *Graph neural networks (GNNs)* are then leveraged for 2D molecular graph representation, based on the *message-passing neural network (MPNN)* framework.

In general, a *2D molecular graph* [1] can be represented as $G = (X, (I, E))$ where $X \in R^{N \times d}$ is the *node feature matrix*, and $(I, E)$ is the *sparse adjacency tuple*. $I \in N^{2 \times U}$ encodes *edge indices* in coordinate (COO) format and $E \in R^{U \times s}$ is the *edge feature matrix*. (N is the number of nodes, U is the number of edges, d is the number of node features, and s is the number of edge features.)

## 2.2 Molecular Geometry

*Molecular geometry* [2] is the 3D arrangement of the *atoms* (nodes) and *bonds* (edges) in a molecule. It influences important properties of a molecule including its reactivity and biological activity. 3D Molecular geometry is commonly represented as *3D graphs*.

The *geometry* of 3D graphs [2] is the 3D arrangement of the *nodes* and *edges* in a graph. It is often specified in terms of the *Cartesian coordinates of nodes*. However, such specification depends on the (arbitrary) choice of origin and is too general for specifying geometry. For most 3D graphs, their geometry can be fully specified in terms of *edge distances (d), angles (θ)* and *dihedrals (φ)*. The edge distance is the distance between two nodes connected together. The angle is the angle formed between three nodes across two edges. For three edges connected in a chain, the dihedral is the angle between the plane formed by the first two edges and the plane formed by the last two edges. These are illustrated in the following diagram:

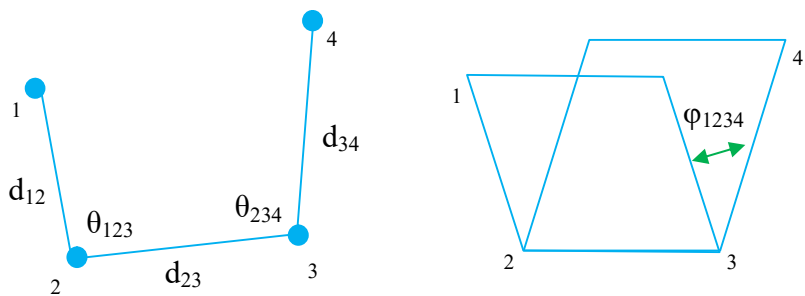

A key advantage of using edge distances, angles and dihedrals to specify geometry is its *invariance to rotation and translation* of the graph.

Specifically, 3D molecular geometry [2] can be specified in terms of *bond lengths* (i.e., edge distances), *bond angles* (i.e., angles) and *dihedral angles* (i.e., dihedrals). The bond length is the average distance between two atoms bonded together. The bond angle is the angle formed between three atoms across two bonds. For three bonds in a chain, the dihedral angle is the angle between the plane formed by the first two bonds and the plane formed by the last two bonds. These are no different from the general case of 3D graphs. Bond lengths, bond angles and dihedral angles can be calculated from the *Cartesian coordinates of atoms* in a molecule, which are generally expressed in the unit of *angstrom (Å)*.

Predicting an accurate 3D molecular geometry is a crucial task for cheminformatics. The outcomes of prediction include bond lengths, bond angles and dihedral angles, among other things.

### 2.3 Molecular Distance Geometry

*Distance geometry* [2, 4-5], of which *molecular distance geometry* is a special case, refers to a foundation of geometry based on the concept of *distances* instead of those of points and lines or point coordinates. For 3D graphs, distance geometry is the characterization and study of their geometry based only on given values of the *distances* between pairs of nodes. From the perspective of distance geometry, therefore, the geometry of 3D graphs can be equivalently specified in terms of *edge distances (d), angle distances ($d^\theta$) and dihedral distances ($d^\varphi$)*. The angle distance is the distance of the *angle edge ($e^\theta$)*, and the dihedral distance is the distance of the *dihedral edge ($e^\varphi$)*. The angle edge is the unconnected end edge between the end nodes of an angle, and the dihedral edge is the unconnected end edge between the end nodes of a dihedral. (We therefore refer to both angle edges and dihedral edges as *end edges*.) These are illustrated in the following diagram, with dashed lines representing angle edges and dotted lines representing dihedral edges:

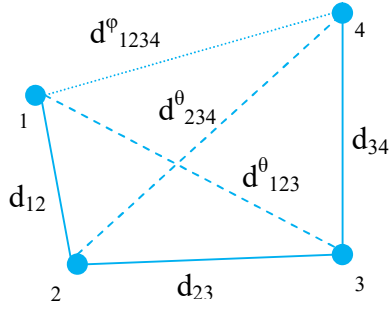

As the case of using edge distances, angles and dihedrals to specify the geometry of 3D graphs, a key advantage [2] of specifying the geometry of 3D graphs in terms of edge distances, angle distances and dihedral distances is its *invariance to rotation and translation*. In addition, it adopts a *unified scheme (distance)* and it reflects *pair-wise node interactions* and their generally local nature, which are additional advantages. These are very useful for *graph convolutions*, which are locally oriented. They are particular useful for *3D molecular geometry*, since electrostatic, intermolecular and other conformation-driven properties of molecules depend on the pair-wise interatomic (internodal) distances.

## 2.4 Geometric Graph Representations

To facilitate the incorporation of *geometry* in deep learning on *3D graphs*, we define three types of *geometric graph representations* [2]: positional, angle-geometric and distance-geometric.

### *Positional Graph Representation*

In general, a *3D graph* can be represented as G = (**X**, (**I**, **E**), **P**) where $\mathbf{X} \in \mathsf{R}^{N \times d}$ is the *node feature matrix*, (**I**, **E**) is the *sparse adjacency tuple*, and $\mathbf{P} \in \mathsf{R}^{N \times 3}$ is the *node position matrix*. $\mathbf{I} \in \mathsf{N}^{2 \times U}$ encodes *edge indices* in coordinate (COO) format and $\mathbf{E} \in \mathsf{R}^{U \times s}$ is the *edge feature matrix*. **P** encodes the Cartesian coordinates of nodes. (N is the number of nodes, U is the number of edges, d is the number of node features, and s is the number of edge features.) This representation is based on node positions; therefore we refer to it as the *positional graph representation*. It depends on the (arbitrary) choice of the origin for the coordinates and it is too general for representing geometry.

### *Angle-Geometric Graph Representation*

From the perspective of *geometry*, a 3D graph can be represented as G = (**X**, (**I**, **E**), (**D**, **Θ**, **Φ**)) where $\mathbf{D} \in \mathsf{R}^{U \times 1}$ is the *edge distance matrix*, $\mathbf{\Theta} \in \mathsf{R}^{U^\theta \times 1}$ is the *angle matrix*, and $\mathbf{\Phi} \in \mathsf{R}^{U^\varphi \times 1}$ is the *dihedral matrix*. (U is the number of edges, $U^\theta$ is the number of angles, and $U^\varphi$ is the number of dihedrals.) This representation centers on (edge distances,) angles and dihedrals; therefore we refer to it as the *angle-geometric graph representation*. It is *invariant to rotation and translation* of the graph, which is a major advantage over the positional graph representation.

### *Distance-Geometric Graph Representation (DG-GR)*

From the perspective of *distance geometry*, a 3D graph can be represented as $G = (X, (I, E), (D, D^\theta, D^\varphi))$ where $D \in R^{U \times 1}$ is the *edge distance matrix*, $D^\theta \in R^{U^\theta \times 1}$ is the *angle distance matrix*, and $D^\varphi \in R^{U^\varphi \times 1}$ is the *dihedral distance matrix*. (U is the number of edges, $U^\theta$ is the number of angles, and $U^\varphi$ is the number of dihedrals.) This representation is based on distances; therefore we refer to it as the *distance-geometric graph representation*. It is *invariant to rotation and translation* of the graph, same as the angle-geometric graph representation. In addition, it reflects *pair-wise node interactions* and their generally local nature, particularly relevant for 3D molecular geometry.

## 3 Geometric Graph Convolutions and DG-GCN

As in [2], we start with *graph neural networks* (*GNNs*) that employ the following *message passing* scheme for node i at a layer:

$$\mathbf{x}_i^* = \gamma(\mathbf{x}_i, \Pi_j \lambda(\mathbf{x}_i, \mathbf{x}_j, \mathbf{e}_{ij}))$$

where $j \in N(i)$ denotes a neighbor node of node i. $\mathbf{x}_i$ is the node feature vector and $\mathbf{e}_{ij}$ is the edge feature vector. $\gamma$ and $\lambda$ denote differentiable update and message functions, respectively, and $\Pi$ denotes a differentiable aggregation function.

The *standard GCN* [7] implements message passing using the adjacency matrix $\mathbf{A}$:

$$\mathbf{X}^* = \check{\mathbf{D}}^{-1/2} \check{\mathbf{A}} \check{\mathbf{D}}^{-1/2} \mathbf{X} \mathbf{W},$$

where $\check{\mathbf{A}} = \mathbf{A} + \mathbf{I}$ denotes the adjacency matrix with inserted self-loops, and $\check{D}_{ii} = \sum_{j=0} \check{A}_{ij}$ its diagonal degree matrix. $A_{ij}$ is one when there is an edge from node i to node j, and zero when there is no edge. $\mathbf{W}$ is the layer-specific weight matrix.

In the context of message-passing GNNs, the standard GCN takes the following shape:

$$\mathbf{x}_i^* = \frac{1}{c_i}(\mathbf{x}_i + \Sigma_j \mathbf{x}_j)\mathbf{W}$$

where $c_i$ is a node-specific normalization constant.

In case the graph has edge weight, $w_{ij}$, the above equation can be expanded as:

$$\mathbf{x}_i^* = \frac{1}{c_i}(\mathbf{x}_i + \Sigma_j\ w_{ij}\mathbf{x}_j)\mathbf{W}$$

This is used by *geometric graph convolutions* [2] with $w_{ij}$ determined from edge distance and N(i) expanded as N(i) = $\mathcal{N}$(i) + $\mathcal{N}^\theta$(i) + $\mathcal{N}^\varphi$(i), where $\mathcal{N}$(i) being the *first-order (1st) "connected"* neighbors, $\mathcal{N}^\theta$(i) the *second-order (2nd) "angle"* neighbors and $\mathcal{N}^\varphi$(i) the *third-order (3rd) "dihedral"* neighbors.

*Geometric graph convolutions*, in essence, use DG-GR and employ parameterized *edge weight / edge distance power laws*. The combination enables the incorporation of geometry in graph convolutions utilizing *standard GCNs* by (1) expanding the kinds of edges involved to include not just *(connected) edges (e)* with neighbor nodes, but also *angle edges ($e^\theta$)* with second-order-neighbor nodes and *dihedral edges ($e^\varphi$)* with third-order-neighbor nodes, and (2) assigning different *weight*s to different edges based on their kind and distance. The parameters in the power laws can be fixed using *reference values* or determined using *Bayesian hyperparameter optimization*.

*DG-GCN (distance-geometric graph convolutional network)* [3], a message-passing graph convolutional network based on DG-GR, improves upon this. Similar to geometric graph convolutions, DG-GCN considers all edges that are important to 3D graph geometry in graph convolutions. These include *(connected) edges*, *angle edges* and *dihedral edges*. However, instead of standard GCN layers, it utilizes *continuous-filter convolutional layers (CFConv)*, with *filter-generating networks*, which enable learning of filter weights from edge distances. This avoids the need for hand-crafted edge weight / edge distance power laws.

## 4 Distance-Geometric Graph Attention Network (DG-GAT)

In this work, to make further improvement, we adopt the new *graph attention network with dynamic attention (GATv2)* [8-9] for use with DG-GR and propose the 3D *distance-geometric graph attention network (DG-GAT)*. Essentially, we replace CFConv in DG-GCN (or standard GCN in geometric graph convolutions) with GATv2 as the convolution layer in DG-GAT. GATv2 is a great fit for DG-GR since the attention can vary by node and by distance between nodes. Further, GATv2 is simpler to understand and use than CFConv, and it is more general and more powerful.

## 4.1 GATv2

In *GATv2*, every node can attend to any other node:

$$x_i^* = \sigma(\alpha_{i,i}Wx_i + \Sigma_j \alpha_{i,j}Wx_j)$$

where j ∈ N(i) denotes a neighbor node of node i, $\alpha_{i,j}$ is the *attention coefficient*, and σ is a nonlinearity. The attention function is defined as:

$$\alpha_{i,j} = \exp(e(x_i, x_j)) / \Sigma_k \exp(e(x_i, x_k)),$$

where

$$e(x_i, x_j) = a^T \text{LeakyReLU}(W[x_i \| x_j]).$$

**a** and **W** are learned and ‖ denotes vector concatenation.

If the graph has edge features $e_{ij}$, $e(x_i, x_j)$ is replaced by

$$e(x_i, x_j, e_{ij}) = a^T \text{LeakyReLU}(W[x_i \| x_j \| e_{ij}]).$$

This is used by *DG-GAT* with *edge distances* as edge features and N(i) expanded as N(i) = $\mathcal{N}(i) + \mathcal{N}^\theta(i) + \mathcal{N}^\varphi(i)$, with $\mathcal{N}(i)$ being the *first-order (1st) "connected"* neighbors, $\mathcal{N}^\theta(i)$ the *second-order (2nd) "angle"* neighbors and $\mathcal{N}^\varphi(i)$ the *third-order (3rd) "dihedral"* neighbors. Specifically, $e_{ij}$ is determined from $d_{ij}$ using [3]

$$e_{ij} = 0.5 * \cos((d_{ij} / d_{cutoff}) * \pi) + 1),$$

where $d_{cutoff}$ is the distance cutoff. $e_{ij}$ varies from 1 ($d_{ij} = 0$) to 0 ($d_{ij} = d_{cutoff}$).

*DG-GAT*, in essence, uses DG-GR and employs *cosine-transformed edge distances as edge features*. The combination enables the incorporation of geometry in graph convolutions utilizing *GATv2* by (1) expanding the kinds of edges involved to include not just *(connected) edges (e)* with neighbor nodes, but also *angle edges ($e^\theta$)* with second-order-neighbor nodes and *dihedral edges ($e^{\varphi}$)* with third-order-neighbor nodes, and (2) using both *node features and edge features* for computing attention coefficients. (Note that additional edge features can be used, beyond edge distances, if needed.)

## 4.2 DG-GAT

As stated previously, *DG-GAT* considers all edges that are important to graph geometry in graph convolutions. These include *(connected) edges (e)* with $1^{st}$ neighbors, *angle edges ($e^{\theta}$)* with $2^{nd}$ neighbors, and *dihedral edges ($e^{\varphi}$)* with $3^{rd}$ neighbors. The following diagram [12] shows the neighborhood of a node of a 3D molecular graph where case a) includes $1^{st}$ neighbors (black), case b) includes $1^{st}$ neighbors and $2^{nd}$ neighbors (blue), and case c) includes $1^{st}$ neighbors, $2^{nd}$ neighbors and $3^{rd}$ neighbors (red). It can be seen that DG-GAT, i.e., case c), fully captures the *local geometry (edge distances, angles, and dihedrals) and substructures* of a node in graph convolutions.

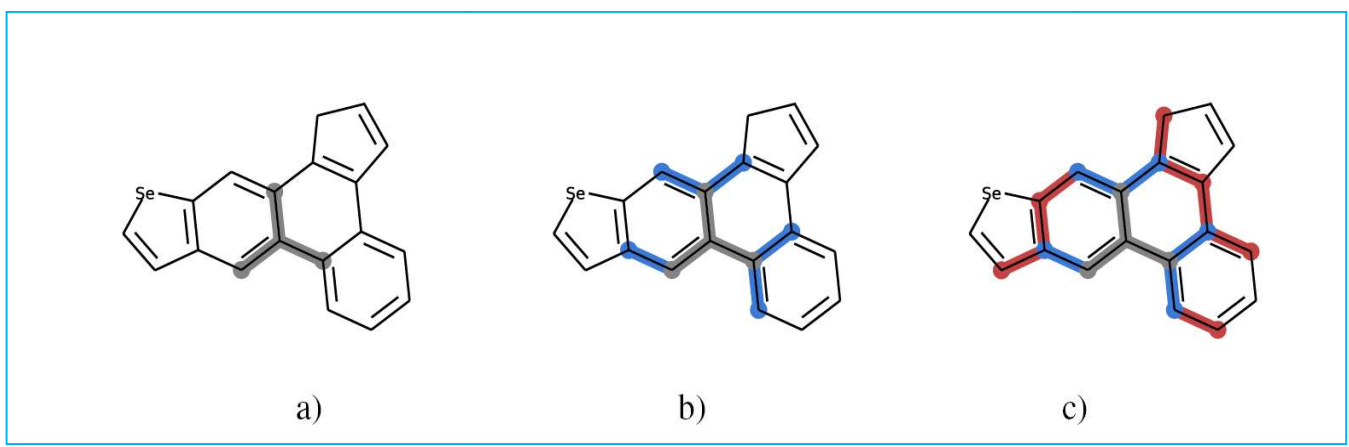

a)    b)    c)

The architecture of DG-GAT is shown below, where $\{d_{ij}\}$ denotes the set of all distances, including edge distances (d), angle distances ($d^{\theta}$) and dihedral distances ($d^{\varphi}$). Note that, as discussed previously, all distances are cosine-transformed and used as edge features.

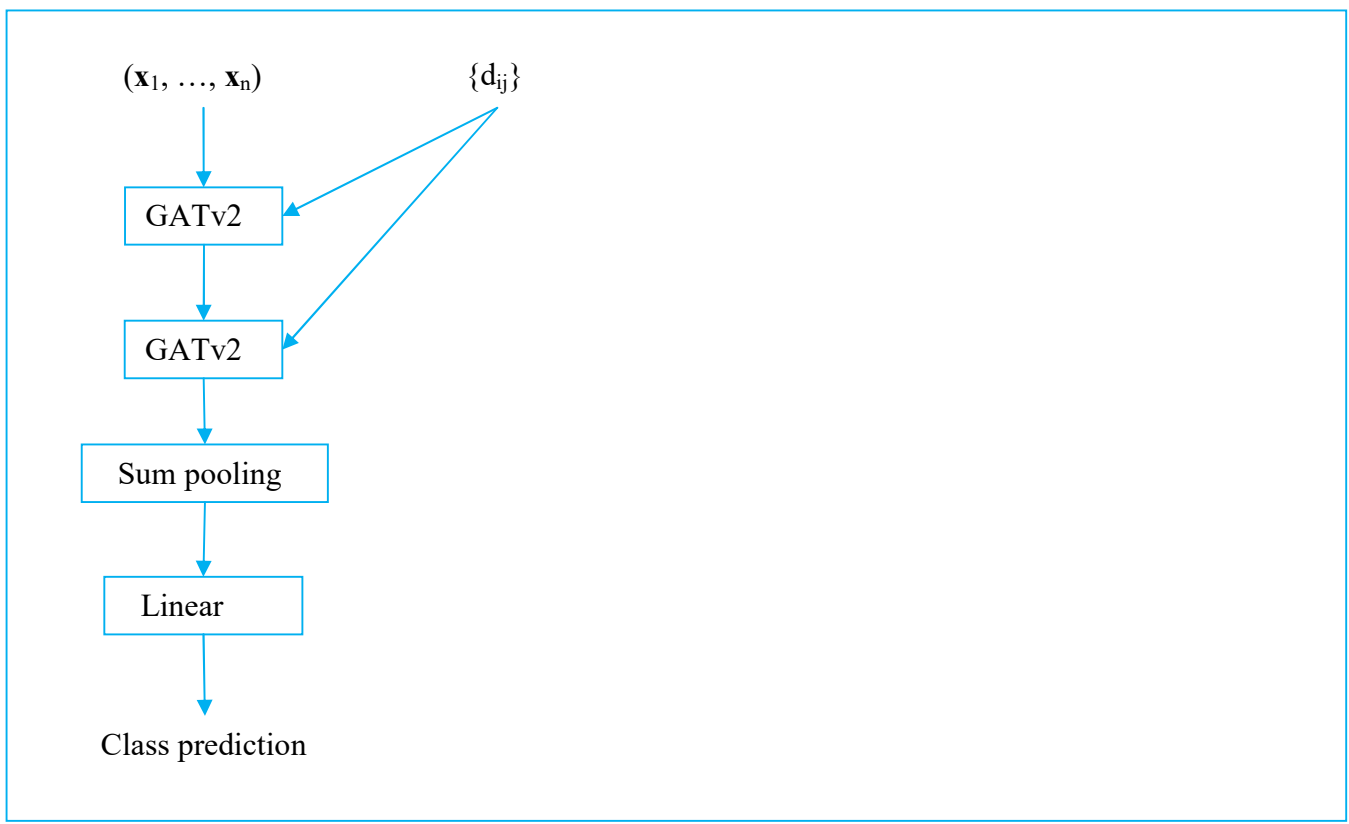

DG-GAT is implemented using *PyTorch Geometric (PyG)* [10]. In particular, the implementation adopts *GATv2Conv* as the convolutional layer, with the attention coefficients computed based on both *node features and edge features*.

## 5 Experiments

For experiments, we use the *ESOL* and *FreeSolv* datasets provided by geo-GCN [11] as well as the *QM9* dataset in PyG, which contain molecular graph data including *3D node coordinates*.

### 5.1 ESOL and FreeSolv

A number of experiments have been carried out using the *ESOL* and *FreeSolv* datasets. Specifically, dataset files provided by Geo-GCN are used, same as [2-3]. These are small datasets with *901 / 113 / 113* and *510 / 65 / 64* training / test / validation samples, respectively. However, they are sufficient for our purpose since our focus is on qualitatively comparing results of DG-GAT with those of (standard) GCN and DG-GCN, all based on the same sample sizes. That is, our interest is on relative accuracy not absolute accuracy.

As in [2-3], we first consider three cases of DG-GAT: edges ($1^{st}$ *Nbrs*) which include $1^{st}$ neighbor nodes, edges + angle edges ($2^{nd}$ *Nbrs*) which include $1^{st}$ and $2^{nd}$ neighbor nodes, and edges + angle edges + dihedral edges ($3^{rd}$ *Nbrs*) which include

1st, 2nd and 3rd neighbor nodes. We include the 1st Nbrs and 2nd Nbrs cases to verify and show the consistency of DG-GAT results. The (full) DG-GAT results are represented by the *3rd Nbrs* cases. These are shown in the following table:

| Dataset | Model | RMSE |
|---|---|---|
| ESOL | DG-GAT – 1st Nbrs | 0.3741 |
| | DG-GAT – 2nd Nbrs | 0.3362 |
| | DG-GAT – 3rd Nbrs | 0.3133 |
| FreeSolv | DG-GAT – 1st Nbrs | 0.2671 |
| | DG-GAT – 2nd Nbrs | 0.2529 |
| | DG-GAT – 3rd Nbrs | 0.2599 |

The *DG-GAT* results are compared to those of GCN and DG-GCN in the following table. It can be seen that they show improvement over both GCN and DG-GCN. The DG-GAT results are consistent with, but better than, those of DG-GCN. On the other hand, the DG-GAT results show major improvement (*31% and 38%,* respectively, for ESOL and FreeSolv) over those of GCN based on 2D molecular graphs. This demonstrates the utility and value of DG-GAT for *deep learning based on 3D molecular geometry*.

| Dataset | Model | RMSE |
|---|---|---|
| ESOL | GCN | 0.4573 |
| | DG-GCN | 0.3231 |
| | DG-GAT | 0.3133 |
| FreeSolv | GCN | 0.4183 |
| | DG-GCN | 0.2890 |
| | DG-GAT | 0.2599 |

## 5.2 QM9

The DG-GAT results are shown in the following table, and compared to those of GCN and DG-GCN. It can be seen that among the twelve properties, the following five properties (marked with *) exhibit significant geometric effect (in comparison to GCN based on 2D molecular graphs): mu (-44%), LUMO (-45%), gap (-21%), $R^2$ (-42%) and Cv (-42%). The DG-GAT results are consistent with, but better than, those of DG-GCN. For other properties, the results are inconclusive due to the small sample size (1100 / 100 / 100) used and, therefore, the likelihood of overfitting.

| QM9 Molecular Property | Model | RMSE (Relative) |
|---|---|---|
| mu (dipole moment) * | GCN | 0.8965 |
| | DG-GCN | 0.6425 |
| | DG-GAT | 0.5023 |
| alpha (isotropic polarizability) | GCN | 0.0257 |
| | DG-GCN | 0.0219 |
| | DG-GAT | 0.0309 |
| HOMO (highest occupied molecular orbital energy) | GCN | 0.0596 |
| | DG-GCN | 0.0646 |
| | DG-GAT | 0.0503 |
| LUMO (lowest unoccupied molecular orbital energy)* | GCN | 1.7978 |
| | DG-GCN | 1.4570 |
| | DG-GAT | 0.9818 |
| gap (gap between HOMO and LUMO)* | GCN | 0.1076 |
| | DG-GCN | 0.0863 |
| | DG-GAT | 0.0848 |
| $R^2$ (electronic spatial extent)* | GCN | 0.1601 |
| | DG-GCN | 0.0925 |
| | DG-GAT | 0.0928 |
| ZPVE (zero point vibrational energy) | GCN | 0.0101 |
| | DG-GCN | 0.0106 |
| | DG-GAT | 0.0080 |
| $U_0$ (internal energy at 0K) | GCN | 0.0095 |

|  | DG-GCN | 0.0262 |
|---|---|---|
|  | DG-GAT | 0.0002 |
| U (internal energy at 298.15K) | GCN | 0.0095 |
|  | DG-GCN | 0.0262 |
|  | DG-GAT | 0.0004 |
| H (enthalpy at 298.15K) | GCN | 0.0095 |
|  | DG-GCN | 0.0262 |
|  | DG-GAT | 0.0003 |
| G (free energy at 298.15K) | GCN | 0.0095 |
|  | DG-GCN | 0.0262 |
|  | DG-GAT | 0.0005 |
| Cv (heat capacity at 298.15K)* | GCN | 0.0416 |
|  | DG-GCN | 0.0250 |
|  | DG-GAT | 0.0240 |

## 5.3 Note

Note that the experiments show *DG-GAT* generally has better results than DG-GCN. Further, as mentioned previously, *GATv2* is simpler to understand and use than CFConv, and it is more general and more powerful. Therefore, DG-GAT is preferred over DG-GCN to use for *deep learning based on 3D molecular geometry*.

## 6 Conclusion

To facilitate the incorporation of 3D molecular geometry in deep learning for molecular science, we adopt the new graph attention network with dynamic attention (GATv2) for use with the 3D distance-geometric graph representation (DG-GR) and propose the 3D distance-geometric graph attention network (DG-GAT). Experimental results of DG-GAT for the ESOL and FreeSolv datasets show major improvement (31% and 38%, respectively) over those of the standard graph convolution network based on 2D molecular graphs. The same is true for the QM9 dataset.

Our work demonstrates the utility and value of DG-GAT for deep learning based on 3D molecular geometry.

**Acknowledgement:** Thanks to my wife Hedy (郑期芳) for her support.